\documentclass[review]{fcs}
\usepackage{times}
\usepackage{soul}
\usepackage{url}
\usepackage{graphicx}
\usepackage{amsmath}
\usepackage{amsthm}
\usepackage{booktabs}
\usepackage{algorithm}
\usepackage{algorithmic}
\usepackage[switch]{lineno}
\usepackage{multirow}
\usepackage{tabularx}
\usepackage{colortbl}

\usepackage{amssymb} 
\usepackage{cleveref} 
\usepackage{listofitems}
\usepackage{mathtools}
\usepackage{enumitem}
\usepackage{listings}
\usepackage{booktabs}
\usepackage{wrapfig}

\definecolor{lstbackground}{rgb}{0.95,0.95,0.95}
\definecolor{lstcomment}{rgb}{0.5,0.5,0.5}
\definecolor{lstkeyword}{rgb}{0,0.5,0.5}
\definecolor{lststring}{rgb}{0.5,0,0}

\newcommand{\rt}[1]{\colorbox{pink}{#1}}
\newcommand{\gt}[1]{\colorbox{green}{#1}}

\lstdefinestyle{academic}{
    language=tex,
    basicstyle=\ttfamily\footnotesize,
    backgroundcolor=\color{lstbackground},
    commentstyle=\color{lstcomment},
    stringstyle=\color{lststring},
    showstringspaces=false,
    breaklines=true,
    frame=single,
    framesep=5pt,
    xleftmargin=10pt,
    xrightmargin=10pt,
    tabsize=4,
    captionpos=b,
    breakindent=0pt,
    rulecolor=\color{black},
    escapeinside={(*@}{@*)},
}
\definecolor{codegreen}{RGB}{86, 156, 74}
\definecolor{codegray}{RGB}{105, 105, 105}
\definecolor{codepurple}{RGB}{161, 21, 132}
\definecolor{backcolour}{RGB}{240, 240, 240}

\lstdefinestyle{pythonstyle}{
    language=Python,
    backgroundcolor=\color[RGB]{255, 255, 255, 0},
    commentstyle=\color{codegray},
    keywordstyle=\color{codepurple},
    numberstyle=\tiny\color{codegray},
    stringstyle=\color{codegreen},
    basicstyle=\ttfamily\tiny,
    breakatwhitespace=false,
    breaklines=true,
    captionpos=b,
    keepspaces=true,
    numbersep=5pt,
    showspaces=false,
    showstringspaces=false,
    showtabs=false,
    tabsize=4,
    frame=single,
    framexleftmargin=10pt,
    framexrightmargin=10pt,
    framextopmargin=6pt,
    framexbottommargin=6pt,
    rulecolor=\color{backcolour},
    aboveskip=12pt,
    belowskip=12pt,
}

\newcommand\msmfmt{00.00}
\newcommand\mssep{\ensuremath{_\pm}}
\newcommand\mssfmt{_{00.00}}
\newcommand\msstdalign{l}

\newcommand{\basicmeanstyle}[1]{#1}%
\newcommand{\basicstdstyle}[1]{#1}%
\newsavebox\CBox 
\newcommand*\bfmeanstyle[1]{\sbox\CBox{\ensuremath{\basicmeanstyle{#1}}}\resizebox{\wd\CBox}{\ht\CBox}{\ensuremath{\mathbf{\basicmeanstyle{#1}}}}}%
\newcommand*\bfstdstyle[1]{\sbox\CBox{\ensuremath{_{\basicstdstyle{#1}}}}\resizebox{\wd\CBox}{\ht\CBox}{\ensuremath{\mathbf{_{\basicstdstyle{#1}}}}}}%

\newlength{\meanlen}
\newlength{\stdlen}
\newcommand{\meanstyle}[1]{\basicstdstyle{#1}}%
\newcommand{\stdstyle}[1]{\basicstdstyle{#1}}%

\newcommand{\msprintmean}[1]{\ensuremath{\mathmakebox[\meanlen][r]{\meanstyle{#1}}}}
\newcommand{\msprintstd}[1]{\mssep\ensuremath{\mathmakebox[\stdlen][\msstdalign]{_{\stdstyle{#1}}}}}

\newcommand{\setcellformat}[4][l]{
    \renewcommand\msstdalign{#1}%
    \renewcommand\msmfmt{#2}%
    \renewcommand\mssep{#3}%
    \renewcommand\mssfmt{_{#4}}%
    \settowidth{\meanlen}{\ensuremath{\msmfmt}}%
    \settowidth{\stdlen}{\ensuremath{\mssfmt}}%
}

\newcommand{\msc}[2][n]{%
    \let\meanstyle\basicmeanstyle%
    \let\stdstyle\basicstdstyle%
    \ifx b#1\let\meanstyle\bfmeanstyle\let\stdstyle\bfstdstyle\fi%
	\setsepchar{,}%
    \reademptyitems%
	\readlist*\theparam{#2}%
    \ignoreemptyitems%
	\readlist*\theneparam{#2}%
    \msprintmean{\theparam[1]}%
    \ifnum \theparamlen=2 
        \ifnum \theneparamlen=2
            \msprintstd{\theparam[2]}%
        \else
            \phantom{\msprintstd{0}}%
        \fi
    \fi
}

\newcommand{\msb}[1]{\msc[b]{#1}}

\renewcommand{\basicstdstyle}[1]{{\color{red} +#1\%}}
\setcellformat[l]{00.0}{}{0000.0}

\title{Top Pass: Improve Code Generation by Pass@k-Maximized Code Ranking}
\author[1,2]{Zhi-Cun Lyu}
\author[1,2]{Xin-Ye Li}
\author[1]{Zheng Xie}
\author*[1,2]{Ming Li}

\address[1]{National Key Laboratory for Novel Software Technology,\\ Nanjing University, Nanjing 210023, China}
\address[2]{School of Artificial Intelligence, Nanjing University, Nanjing 210023, China}
\fcssetup{
  received       = {April 28, 2024},
  accepted       = {July 04, 2024},
  corr-email     = {lim@nju.edu.cn},
}
\begin{abstract}
Code generation has been greatly enhanced by the profound advancements in Large Language Models (LLMs) recently.
Nevertheless, such LLM-based code generation approaches still struggle to generate error-free code in a few tries when faced with complex problems. To address this, the prevailing strategy is to sample a huge number of candidate programs, with the hope of any one in them could work. 
However, users of code generation systems usually expect to find a correct program by reviewing or testing only a small number of code candidates. Otherwise, the system would be unhelpful. 
In this paper, we propose Top Pass, a code ranking approach that identifies potential correct solutions from a large number of candidates. Top Pass directly optimizes the pass@k loss function, enhancing the quality at the top of the candidate list. This enables the user to find the correct solution within as few tries as possible.
Experimental results on four benchmarks indicate that our Top Pass method enhances the usability of code generation models by producing better ranking results, particularly achieving a 32.9\% relative improvement in  pass@1 on CodeContests  when compared to the state-of-the-art ranking method.
\end{abstract}

\begin{document}

\section{Introduction}

Code generation aims to generate programs automatically based on the requirements described in natural language. 
It is important as programmers could save great time and efforts if correct programs are generated automatically.  
Large Language Models (LLMs) such as ChatGPT have brought the breakthroughs to the field of code generation. 
Specifically, LLMs pre-trained on code corpus substantially enhance their capacities, achieving unprecedented performance in solving problems~\cite{Li2023,Nijkamp2023,Roziere2023,Yaminhu}.
This capacity has served as the foundation for tools like Github Copilot, which capitalizes on the concept of ``natural language programming'' by these LLMs. Previous works have confirmed that these tools improve the coding efficiency of programmers greatly. 

\begin{figure}[t]
    \centering
    \includegraphics[width=\columnwidth]{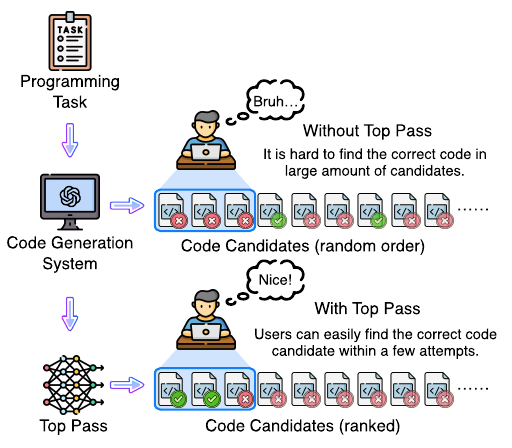}
    \caption{Code generation system with or without Top Pass. The user can only afford testing or reviewing a few code candidates, thus Top Pass enhances the practical value of code generation systems significantly.}
    \label{fig:motivation}
\end{figure}
\begin{figure*}[!h]
    \centering
    \includegraphics[width=\textwidth]{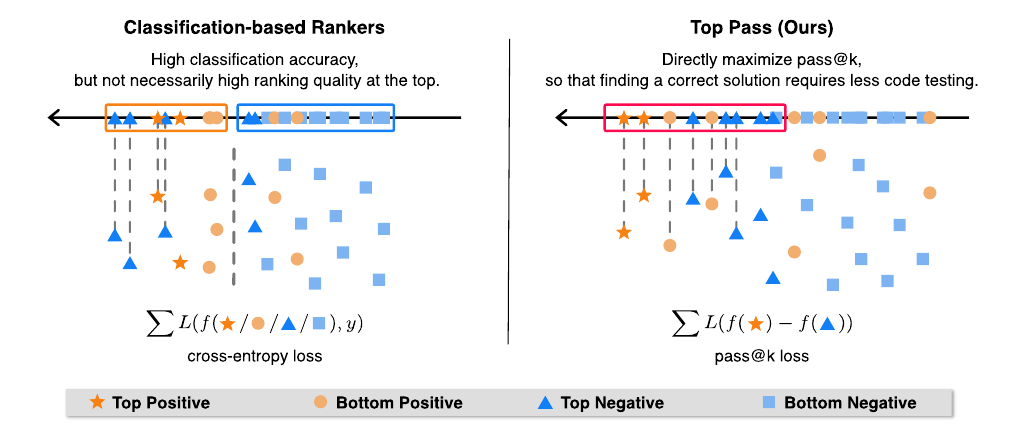} 
    \caption{Top Pass minimizes a novel pass@k loss function that enhances the ranking quality at the top of the code candidate list, so that the user can solve the programming task with fewer attempts.}
    \label{fig:problem-setting}
    \end{figure*}
    
Despite the rapid improvement in code generation system, when dealing with real-world complex problems, it is still a long way for LLM-based code generation system to generate the correct solution within a few tries. 
To deal with this problem, the prevailing strategy is to sample a huge number of program candidates, usually 100~\cite{Chen2023}, 200~\cite{Li2023}, or even 1000~\cite{Zhang2023algo,Chen2021}, with the hope of any one in them could work.
However, when applying code generation system for improving development efficiency, the correct program should be provided within a small candidate set for the generation system to be helpful. Otherwise, the users will spend too much time on reviewing or testing the wrong code candidates and the time loss will outweigh the gain. Such a goal of generating correct program within as few tries as possible is also reflected in the performance measure of the code generation systems, i.e., (average) pass@k. It measures the possibility of a user can find a correct program by testing k candidates. Typically, the parameter k is set to 1, 5, or at most 10, as it is usually unacceptable to test more than 10 candidates to find the correct program. 
However, ensuring the provision of a correct solution within a small number of candidate programs still remains challenging for current LLM-based code generation system.

Other than fine-tuning the LLM itself for better code generation performance, an effective yet economy way to improve pass@k is to rank the candidates according to their possibilities to be correct. 
CodeRanker~\cite{Inala2022} has demonstrated the effectiveness of such a strategy by training a neural ranker with a binary classification objective to determine which candidates are more likely to be correct. Such an approach improves the performance measures (pass@k) on multiple code generation benchmarks with the help of Codex for generating candidates.
However, it models the code ranking as binary classification and ignores the goal is ranking any positive sample to the top. 
As shown in \cref{fig:problem-setting}, classification based rankers could achieve high classification accuracy but fail to rank positive samples to the top. 
How to effectively minimize user effort in utilizing code generation systems still remains a challenging problem currently.

In this paper, we propose to optimize the metric `pass@k' directly. By its definition, this optimization objective emphasizes that ranker model should rank the first positive program before the k-th negative program. 
However, directly optimizing this formulation faces some challenges.
Solely relying on a single positive sample can disrupt the generalization capabilities of the ranker model. 
Instead our method selects a proportion of high-score positive samples to address the challenge. 
Besides, how to improve `pass@k' for different k simultaneously also remains a challenge. 
We relax negative sample selection to deal with it. 
In practice, it can be observed that such sample selection strategies cause the model to focus more on identifying high-quality code snippets.

To evaluate our approach, we conduct extensive experiments on code generation benchmarks including CodeContests, APPS, MBPP and HumanEval.
We demonstrate the effectiveness of our method on four code generation benchmarks. Experiment results show that our method outperforms the baselines by a large margin. 
Then we further analyze the influence of different hyper parameters. 
Also, we show the robustness of Top Pass against the false positives caused by the weak test cases. 

Our contribution includes:
\begin{itemize}
    \item We claim that optimizing pass@k is beneficial for code ranking in generation systems. 
    Unlike previous studies that focus on leveraging additional information for classification, we emphasize the ranking nature of the problem. 
    It is effective in ensuring that the correct solution appears as early as possible in the final ranked candidate list, thereby reducing the need for testing or reviewing incorrect codes manually. 
    \item We propose Top Pass, a novel approach for ranking generated code candidates based on their probabilities of correctness. By optimizing the pass@k loss, Top Pass effectively identifies high-quality correct codes, leading to improved ranking performance.
    \item Experimental results demonstrate that our approach, Top Pass, substantially enhances the utility of current code generation models, by improving the pass@k performance.  
    Top Pass exhibits notable improvements, elevating the pass@1 of ChatGPT from 2.9\% to 7.3\% and DeepSeek-Coder from 5.2\% to 9.7\% on the CodeContests dataset.
\end{itemize}
The remainder of this paper is organized as follows. We first discuss the related work. Then we introduce the proposed method Top Pass. 
Finally, we show the experiments and conclude the paper.

\section{Related Work}
\subsection{Code Generation}
Recently, LLM-based approaches built upon Transformer architecture~\cite{Vaswani2017} have gained prominence for code generation~\cite{OpenAI2018,Black2021,OpenAI2023,Anil2023,Chowdhery2023}. \cite{Chen2021} introduces Codex, a decoder-only language model fine-tuned on public available code and demonstrates impressive code writing ability. Concurrently, \cite{Wang2021} presents CodeT5, an encoder-decoder architecture model based on T5~\cite{Raffel2020} with incorporation of an identifier aware pre-training task. In addition to the pre-training paradigm, \cite{Li2022} employs reinforcement learning to train AlphaCode and leverages supervision derived from the execution results of test cases. This approach achieves beginner-level performance on the Codeforces platform. 
Open-source code large language models LLMs~\cite{Li2023,Luo2023,Gunasekar2023,Roziere2023,bi2024deepseek,Zheng2023a,fried2023incoder} also make significant strides in performance on code generation benchmarks.
Other than fine-tuning code LLM, 
\cite{Chen2023a} enforce the code LLM perform rubber duck debugging by explaining the generated code through few-shot demonstrations. 
Generating the extra test cases is validated useful in practice~\cite{Liu2023,Chen2023,Deng2023}.

\subsection{Code Understanding}
Code ranking heavily depends on good code understanding capacities. 
Typical code understanding tasks include functionality classification, clone detection and code search~\cite{Wang2020,Gu2021,Arakelyan2022,zhongli}. 
Prior works~\cite{Kanade2020,Feng2020,Guo2021,Guo2022,Ahmad2021,wang2021syncobert} focus on enhancing code representation. They treat code as sequence of tokens and use pre-trained langugage models to capture the representation for the sequence. 
A notable work in this domain is CodeBERT~\cite{Feng2020}, which uses encoder-only architecture and utilizes replace token detection objective~\cite{Clark2020} for modeling both natural language and programming language.
\cite{Guo2021} extends it to GraphCodeBERT by introducing two new structure-aware pre-training tasks
to learn the semantics from source code and data flow. 
Additionally, \cite{Ahmad2021} proposes a one-to-one mapping function which elegantly transforms the Abstract Syntax Trees (AST) into the sequences, thus enabling processing AST information in the same manner as natural language and programming language. 
\cite{Guo2022} proposes a novel cross-modal language model, UniXcoder, which leverages multi-modal information including the AST, the comment and the code fragment to enrich the representation learning during training.

\section{The Top Pass Method}
To alleviate the time cost of users in reviewing or testing the tremendous candidate programs, we propose a novel post-process method named Top Pass focusing on ranking the positive samples to the top. 
We first formalize the problem and then introduce the pass@k loss, which serves as the basis of Top Pass. Last, we describe the overall loss function for training the ranker. 
\begin{figure*}[!t]
    \centering
    \includegraphics[width=2.0\columnwidth]{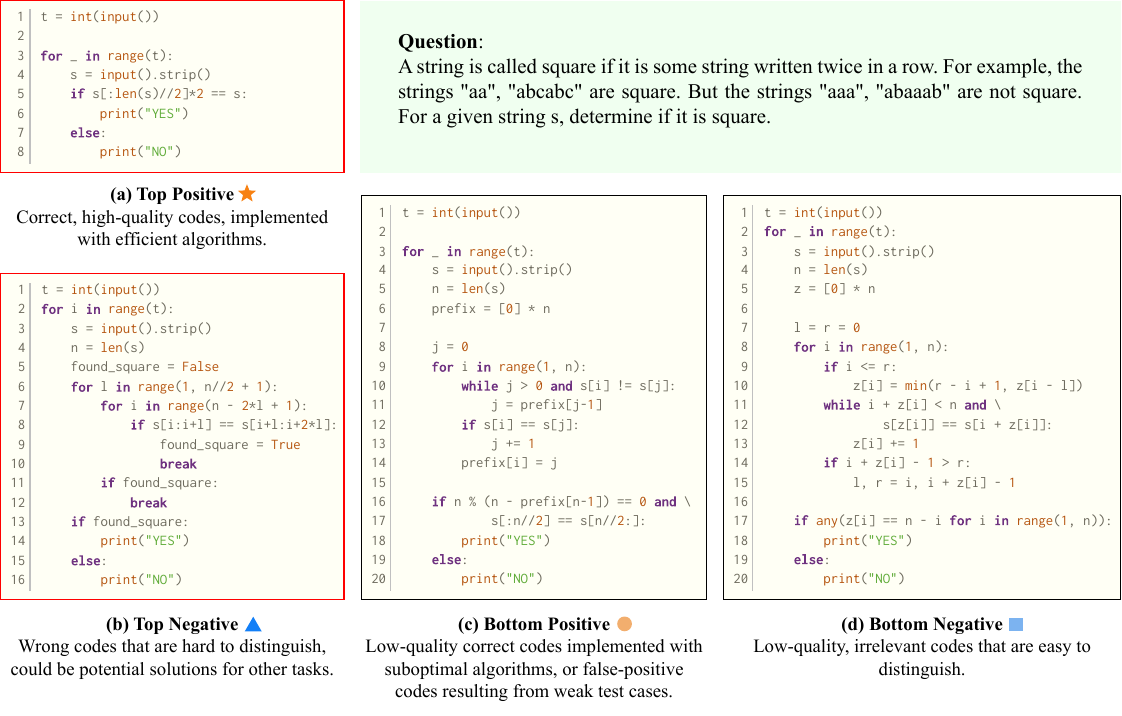}
    \caption{Examples for (a) top positive, (b) top negative, (c) bottom positive, and (d) bottom negative. Pass@k loss gives more significance to the top positive/negative codes, directing the ranking model towards identifying high-quality solutions instead of indistinguishable wrong codes.}
    \label{fig:example}
\end{figure*}

\subsection{Code Generation and Ranking}
For a programming task $Q$ described in natural language, the code generation system is asked to generate $n$ code candidates $\mathcal{C} = \{C_i\}_{i=1}^n$. We use the symbol $y_i = 1$ to denote the candidate is a correct solution (positive), and $y_i = 0$ otherwise (negative). Confirming the correctness of the candidates requires manual review or unit test, thus at most k candidates can be reviewed or tested by the user to find a correct solution. The typical value of k is usually no more than 10, which is significantly smaller than $n$, the total number of candidates. Since the order of the candidates is random, it is doubtful whether the correct solution can be found within k attempts, even if it exists.
Suppose the set of the k tested candidates is $\mathcal{C^*}$, whether the solution of the programming task is found within the k candidates is denoted by `pass@k'.
\begin{definition}[Problem-level pass@k] The problem-level pass@k is defined as if there exist any candidate passes all the test cases in the tested k candidates $\mathcal{C}^*$.
\begin{equation}
    \label{eq:problem-level-pass-at-k}
    \text{pass@k} = \mathbb{I}[\exists C_i \in \mathcal{C}^*, y_i=1]\,.
\end{equation}
\end{definition}
Generally, the \emph{average pass@k} across a group of programming tasks reflects the performance of a code generation system. It indicates how many tasks could be solved if the user can merely afford a maximum of k attempts for each task.

When the code generation system provides the candidates in a random order, the expectation of pass@k can be calculated by averaging pass@k over all k-subsets of $\mathcal{C}$.
\begin{definition}[Expectation of pass@k for random candidate list~\cite{Chen2021}]
The expectation of pass@k can be defined as 
\begin{equation}
    \label{eq:estimated-pass-at-k}
    \text{estimated pass@k} = 
    1-\frac{\binom{n-c}{k}}{\binom{n}{k}}\,,
\end{equation}
where $c$ is the number of the correct candidates.
\end{definition}

It can be inferred from \cref{eq:estimated-pass-at-k} that even if the code generation system provides a correct solution, the likelihood of the user finding it is still low. 
Besides, for a given set of code candidates, the pass@k also fluctuates hugely due to the random order. It yields a variance of 
\begin{equation}
    \label{eq:variance}
    \text{Var} = 
    \left(1-\frac{\binom{n-c}{k}}{\binom{n}{k}}\right) \ \frac{\binom{n-c}{k}}{\binom{n}{k}}\,,
\end{equation}
which makes code generation system unreliable.

To mitigate the unreliability of the code generation system, we introduce code ranking, which involves incorporating a ranker model 
$f$ to discern codes with a higher likelihood of correctness. The ranker model assigns a correctness score to each candidate $s_i=f(Q, C_i)$. By sorting the candidate list based on these scores, the user has a higher probability of finding the correct solution among the k candidates with the highest scores.

\subsection{Pass@k Loss}

As described above, ranking the candidates properly can significantly enhance the performance and reliability of the code generation system. Previous study~\cite{Inala2022} formulates the ranking problem as a binary classification problem. However, such an approach neglects the ranking nature of the task, resulting in an unsatisfactory quality of the selected candidate set, particularly those at the top. 

In this paper, we propose to directly optimize the pass@k, which is the ultimate goal of the code generation systems.
Let $\mathcal{C}_{+} = \{C_{+,1}, C_{+,2}, \cdots\}$ and $\mathcal{C}_{-} = \{C_{-,1}, C_{-,2}, \cdots\}$ denote the sets of positive and negative candidates, respectively, sorted in descending order based on their correctness scores.
Derived from \cref{eq:problem-level-pass-at-k}, pass@k can be reformulated as:
\begin{equation}
    \label{eq:ranked_pass_at_k}
    \text{pass@k}=\mathbb{I}[f(Q,C_{+,1}) > f(Q,C_{-,k})]\,.
\end{equation}
That is, the model should rank the first positive program before the k-th negative program.
Based on the above formulation, the learning problem of ranking the code candidates can be immediately formulated as:
\begin{equation}
    \min_{f} \quad 1-\mathbb{I}[f(Q,C_{+,1}) > f(Q,C_{-,k})]\,.
\end{equation}
This optimization problem exactly reflects the objective of `pass@k' metric, which involves \emph{ranking any one of the positive candidate codes to the top k of the list}. However, directly solving the above maximization problem encounters several practical challenges. Next, we will describe how to address these challenges.

\subsubsection{Identifying top positive programs}
The above formulation only requires the model to identify any one of the correct programs. However, due to the randomness of the candidates, training the model with only one positive program diminishes the model's ability to identify the correct code and consequently lowers its generalization performance. In practice, we relax the limitation of using only one correct program with the maximum score and allow the use of multiple correct programs with top scores. 
Let $\mathcal{C}_{+}^p = \{C_{+,1}, C_{+,2}, \cdots, C_{+,m}\}$, where we select top proportion $p$ positive candidates with highest confidence. 
We introduce a hyper-parameter to filter certain percentage of candidates. Ideally, the positive snippets with higher scores are the ones that are easily identified and are potentially the snippets written with commonly used algorithms and high-quality codes. On the other hand, the positive snippets with lower scores are those that are difficult to identify, often implemented with sub-optimal algorithms, low-quality codes, or even false positive snippets due to weak test cases. For examples, in the case described in \cref{fig:example}, the top positive program is correct and in elegant implementation. 
\subsubsection{Identifying top negative programs}
Instead of using a single k-th top negative snippet for a specific k, we opt to use a portion of the top negative candidates as the negative set.
Let $\mathcal{C}_{-}^q = \{C_{-,1}, C_{-,2}, \cdots, C_{-,n}\}$, where we select top proportion $q$ negative candidates with highest score.
The proportion is usually larger than the typical value of k, allowing the ranker model to maximize pass@k for different common k values simultaneously. The selected negative programs with high scores are typically those that are challenging to distinguish from the positive programs. These programs are also typically high-quality codes but do not solve the given problem. Training with these negative programs enhances the discriminative abilities of the ranker model. On the other hand, the negative ones with lower scores are mostly irrelevant or low quality codes that are easily identified. By not training with those programs, we prevent the model from excessively focusing on irrelevant codes. For example, in the case described in \cref{fig:example}, the problem aims to implement python program to check whether the input string has the ``square'' property, which means the string could be divided into two same substrings. The top negative programs implements an algorithm to check whether there exists a substring satisfying the ``square'' property in the given input string. It is flawless and easy to read, but it misunderstand the requirements.

By combining the aforementioned techniques, we propose our loss function for maximizing pass@k with the following formulation:
\begin{equation}
    \sum_{C^+_i \in \mathcal{C}_{+}^p} \sum_{C^-_j \in \mathcal{C}_{-}^q} \mathbb{I}[f(Q,C^+_i)>f(Q,C^-_j)]\,.
\end{equation}
It can be equivalently rewritten with the 0-1 loss:
\begin{equation}
    \sum_{C^+_i \in \mathcal{C}_{+}^p} \sum_{C^-_j \in \mathcal{C}_{-}^q} l_{01}(f(Q,C^+_i)-f(Q,C^-_j))\,,
\end{equation}
where \(l_{01}(z) = \frac{1}{2}(1-sign(z))\).
In practice, we use hinge square as a convex and smooth surrogate loss function for better optimization:
\begin{equation}
L_{pass@k} = \sum_{C^+_i \in \mathcal{C}_{+}^p} \sum_{C^-_j \in \mathcal{C}_{-}^q} l(f(Q,C^+_i)-f(Q,C^-_j))\,,
\label{eq:loss_pass_at_k}
\end{equation}
where the surrogate loss \(l(z) = (1-z)^2\).

The fine-tuning of the ranker model is based on the minimization of the following loss function: 
\begin{equation}
    \label{eq:loss-function}
    L = L_{pass@k} + \lambda L_{cls}\,,
\end{equation}
where $L_{pass@k}$ is our pass@k loss described previously, $L_{cls}$ is a standard cross-entropy loss, and $\lambda$ is a coefficient of $L_{cls}$, typically set to 0.3.


\section{Experiments}
\begin{table*}[!ht]
\renewcommand{\basicstdstyle}[1]{{\color{red} +#1\%}}
\setcellformat[l]{00.0}{}{0000.0}
\newcommand{\cc}{\cellcolor{gray!20}}
\centering
\small
\begin{tabular}[\textwidth]{llllllll}
\toprule
\textbf{Category}                  & \textbf{Method}              & \textbf{Pass@1}   & \textbf{Pass@2}    & \textbf{Pass@3}    & \textbf{Pass@5}   & \textbf{Pass@10}\\ \midrule
\multirow{7}{*}{Standalone LLM}    & PG-TD                        & \msc{0.7,}        & \msc{1.1,}         & \msc{-,}           & \msc{-,}          & \msc{2.5,}\\
                                   & Codex                        & \msc{0.7,}        & \msc{1.2,}         & \msc{-,}           & \msc{-,}          & \msc{3.0,}\\
                                   & WizardCoder                  & \msc{2.0,}        & \msc{-,}           & \msc{-,}           & \msc{3.3,}        & \msc{-,}\\
                                   & WizardCoder + CodeChain      & \msc{2.5,}        & \msc{-,}           & \msc{-,}           & \msc{3.3,}        & \msc{-,}\\
                                   & ChatGPT                      & \msc{2.9,}        & \msc{4.8,}         & \msc{6.3,}         & \msc{8.6,}        & \msc{12.1,}\\
                                   & DeepSeek-Coder               & \msc{5.2,}        & \msc{7.9,}         & \msc{9.8,}         & \msc{12.5,}       & \msc{16.4,}\\ \midrule
\multirow{2}{*}{LLM with testcases}& CodeT                        & \msc{2.1,}        & \msc{2.3,}         & \msc{-,}           & \msc{-,}          & \msc{5.3,}\\
                                   & ALGO                         & \msc{5.6,}        & \msc{5.6,}         & \msc{-,}           & \msc{-,}          & \msc{7.7,}\\ \midrule
\multirow{4}{*}{LLM with ranker}   & CodeRanker w/ ChatGPT        & \msc{6.1,}   & \msc{9.1,}     & \msc{9.7,}     & \msc{10.3,}   & \msc{12.7,}\\
                                   & \cc Top Pass w/ ChatGPT      & \cc\msc{7.3,19.7}& \cc\msc{10.3,13.2}& \cc\msc{12.7,30.9}& \cc\msc{13.3,29.1}& \cc\msc{15.2,19.7}\\
                                   & CodeRanker w/ DeepSeek-Coder  & \msc{7.3,}    & \msc{9.1,}     & \msc{10.3,}     & \msc{13.3,}    & \msc{17.0,}\\
                                   & \cc Top Pass w/ DeepSeek-Coder& \cc\msb{9.7,32.9} & \cc\msb{12.7,39.6} & \cc\msb{13.3,29.1} & \cc\msb{14.5,9.0}& \cc\msb{18.2,7.1}\\
\bottomrule
\end{tabular}
\caption{Experiment results on CodeContests dataset. The best results of each metric are presented in \textbf{bold}. and the numbers in \textcolor{red}{red} indicate the relative improvements of pass@k compared to CodeRanker.}
\label{tab:CC-full-testcases}
\end{table*}

We conduct extensive experiments to verify the effectiveness of our Top Pass method. The experiment results shows that (1) Our approach, Top Pass, improves pass@k on four code generation benchmarks. (2) the improvement brought by our Top Pass method is significant across datasets with different false positive rates. (3) the impact of hyper-parameters on the performance of Top Pass is relatively smooth. 

\subsection{Experiment Setup}
\paragraph{Dataset.}
We consider four code generation benchmarks for evaluation. 
(1) \textbf{CodeContests}~\cite{Li2022} is created for fine-tuning and evaluating AlphaCode and consists of competitive programming problems with augmented test cases. The training set contains 13328
tasks, the validation set 117 tasks and the test set 165 tasks. 
(2) \textbf{APPS}~\cite{Hendrycks2021} consists of 5000 training and 5000 test programming tasks collected from popular programming competition platforms such as LeetCode. Programming tasks are split into three difficulty levels, introductory, interview and competition. 
(3) \textbf{MBPP}~\cite{Austin2021} consists of 974 mostly basic python programming tasks, designed to be solvable by entry level programmers. (4) \textbf{HumanEval}~\cite{Chen2021} consists of 164 human-written test tasks and most of the tasks can be solved with a few lines of python code.
On first three benchmarks, we train ranker model on the train split and evaluate it on the test split of the dataset. To measure the transfer ability of our method between two different datasets, we evaluate the pass@k metrics on HumanEval using the ranker model trained on CodeContests without additional training.

We construct the dataset for training the ranker model and testing followings 2 steps. (1) For each problem description, we use LLM to generate 200 candidate python programs. (2) Executing the python candidate programs against test cases provided in the dataset and collect the execution information such as return value and execution time. If the program passes all the test cases within time limit, it is considered correct (positive). Otherwise, it is considered wrong (negative). If the program fail for one test case, we immediately label it as negative and do not execute the left test cases. In this way, we label all candidate programs as positive or negative according to test cases.

\paragraph{Metric.}
We use pass@k (k=1,2,3,5,10) for evaluation. This metric is first computed for each individual problem, then averaged across multiple problems.
The computation of pass@k depends on whether the methods include a ranking step or not. For those methods without a ranking step, we report the expected value of pass@k, as defined in \cref{eq:estimated-pass-at-k}. For methods that do include a ranking step, pass@k is either 0 or 1 for each problem, depending on whether any of the top-k candidates pass all the test cases, as detailed in \cref{eq:ranked_pass_at_k}.

\paragraph{Ranker model.} Our ranker model consists of a pre-trained transformer based encoder and a classification head. We use CodeBERT~\cite{Feng2020} as the base encoder. The input of the model is the concatenation of programming task description and python code, separated by a special token, namely [CLS], $Q$, [SEP], $C$, [EOS]. The hidden representation of the first token [CLS] is used for representing the whole sequence. Subsequently, the classification head is applied to the representation to determine the score $s_i$.

\begin{table*}[!ht]
\setcellformat[l]{00.0}{}{000.0}

\centering
\small
\begin{tabular}{lllllllll}
\toprule
\multirow{2}{*}[-3bp]{\textbf{Method}} & \multicolumn{4}{c}{\textbf{Pass@1}} & \multicolumn{4}{c}{\textbf{Pass@5}} \\
\cmidrule(lr){2-5} \cmidrule(lr){6-9}
                          & \textbf{Intro} & \textbf{Inter} & \textbf{Comp} & \textbf{Total} & \textbf{Intro}& \textbf{Inter}& \textbf{Comp}  & \textbf{Total} \\
\midrule 
Codex                 & \msc{4.1,}     & \msc{0.1,}     & \msc{0.0,}    & \msc{0.9,}     & \msc{9.7,}    & \msc{0.5,}    & \msc{0.1,}     & \msc{2.3,}\\
AlphaCode                 & \msc{-,}       & \msc{-,}       & \msc{-,}      & \msc{-,}       & \msc{14.4,}   & \msc{5.6,}    & \msc{4.6,}     & \msc{7.2,}\\
Code-LLAMA 34B            & \msc{-,}       & \msc{-,}       & \msc{-,}      & \msc{-,}       & \msc{32.8,}   & \msc{8.8,}    & \msc{2.9,}     & \msc{12.4,}\\
StarCoder                 & \msc{7.3,}     & \msc{6.9,}     & \msc{4.1,}    & \msc{6.4,}     & \msc{-,}      & \msc{-,}      & \msc{-,}       & \msc{-,} \\
WizardCoder               & \msc{26.0,}    & \msc{4.2,}     & \msc{0.8,}    & \msc{7.9,}     & \msc{-,}      & \msc{-,}      & \msc{-,}       & \msc{-,} \\
code-davinci-002          & \msc{19.1,}    & \msc{4.3,}     & \msc{1.0,}    & \msc{6.6,}     & \msc{42.4,}   & \msc{13.1,}   & \msc{4.0,}     & \msc{17.1,}\\
CodeT                     & \msc{34.6,}    & \msc{8.1,}     & \msc{2.2,}    & \msc{12.2,}    & \msc{-,}      & \msc{-,}      & \msc{-,}       & \msc{-,} \\
\midrule
DeepSeek-Coder        & \msc{40.6,}    & \msc{13.8,}    & \msc{4.3,}    & \msc{17.3,}    & \msc{60.0,}   & \msc{27.3,}   & \msc{11.4,}    & \msc{30.7,}\\
CodeRanker                & \msc{44.6,} & \msc{15.0,} & \msc{6.0,}& \msc{19.1,}& \msc{60.1,}& \msc{27.6,}& \msc{13.0,}& \msc{31.2,}\\
\rowcolor{gray!20} Top Pass& \msb{46.6,4.5}& \msb{16.3,8.7}& \msb{6.6,10.0}& \msb{20.4,6.8}& \msb{60.4,0.5}& \msb{28.8,4.3}& \msb{13.3,2.3}& \msb{32.0,2.6}\\
\bottomrule
\end{tabular}

\caption{Experiment results on APPS dataset. The ``Intro'', ``Inter'', ``Comp'' represent introductory, interview and competition-level tasks, respectively, while  ``Total'' encompasses the whole dataset.}
\label{tab:apps}
\end{table*}

\paragraph{Training setup and hyper-parameters.}
The overall process is that we use a code LLM to generate candidates programs and then fine-tune a light-weight neural ranker model with our proposed pass@k loss. 
For sampling code candidates from LLMs, we sample $N=200$ candidates per programming task in four datasets mentioned above since $N=200$ is sufficient to estimate pass@k. We use nucleus sampling with top-$p=0.95$ and temperature $T=0.8$, consistent with the sampling setting used in Codex experiments~\cite{Chen2021}.
For training ranker model, we tuned the hyper parameter $p$ and $q$ on each benchmark, where $p$ represents the proportion of positive and negative sample selection separately. 
In CodeContests we keep top 90\% positives and top 50\% negatives, while in APPS we keep top 70\% positives and top 60\% negatives separately.
For MBPP we keep top 60\% positives and top 70\% negatives separately. 
We calculate the pass@k loss according to \cref{eq:loss_pass_at_k}. 
And we choose $\lambda=0.3$ in \cref{eq:loss-function} to calculate the final loss.
We fine-tuned the ranker models on different datasets. 
We used AdamW with a learning rate of 5e-5 and no weight decay. All the experiments are conducted on A100-40GB GPUs.

\subsection{Baselines}
We compare our method with several baselines:
\begin{itemize}[leftmargin=4mm]
    \item Standalone LLM: We compare Codex~\cite{Chen2021}, ChatGPT~\cite{OpenAI2022}, WizardCoder~\cite{Luo2023}, StarCoder~\cite{Li2023}, AlphaCode~\cite{Li2022}, Code LLaMa~\cite{Roziere2023} and DeepSeek-Coder 33B~\cite{bi2024deepseek}.
    It reflects the average case when sampling from LLMs.
    \item LLM with generated test cases: CodeT~\cite{Chen2023} and ALGO~\cite{Zhang2023algo} generate more test cases to filter candidate programs, leading higher probability of selecting the correct solutions. 
    \item LLM with ranker: CodeRanker~\cite{Inala2022} models the code selection as a binary classification and sort the candidates according to predicted scores. 
    The Coder-Reviewer~\cite{Coder-Reviewer} proposes a novel code ranking method according to the product $p(x|y)p(y|x)$.
\end{itemize}

\subsection{Main Results}
\paragraph{CodeContests.} 

Our experiments are summarized in \cref{tab:CC-full-testcases}, with the best performance highlighted in bold.
Our approach, Top Pass, is highlighted with a shaded background.
The results, as depicted in \cref{tab:CC-full-testcases}, reveal that our approach Top Pass outperforms various baselines over all metric pass@k, including standalone LLMs, LLMs with test cases and prior ranker-based approaches. 
(1) With ChatGPT as generator, Top Pass attains a notable 7.3\% pass@1, 12.7\% pass@3 and 13.3\% pass@5. 
(2) Using the DeepSeek-Coder as generator, our method Top Pass also achieves remarkable improvement over all metrics. It attains a outstanding 9.7\% pass@1, 13.3\% pass@3 and 14.5\% pass@5, bringing a 32.9\% relative improvement in pass@1, 29.1\% in pass@3 compared with CodeRanker. Also, we achieve remarkable nearly twice (1.9$\times$) pass@1 metric compared with standalone DeepSeek-Coder.
In conclusion, our Top Pass method improves the quality of ranking at the top on the CodeContests dataset regardless of the code generation LLM. 
This substantial improvement further enhances the practical usage of code generation system. 
The additional time cost during test includes scoring each candidate program and finding the top-K the candidates with the highest scores. It is relatively small compared with the time cost of LLM inference. On CodeContests the additional time cost for each problem is about 1.8s when we want to select top-5 from 200 candidates. Thus, our method remains promising for practicality in real-world applications.

\paragraph{APPS.}
We evaluate our method on APPS benchmarks. 
The experimental results for APPS are presented in \cref{tab:apps}. 
From the table we can see that our method demonstrate a significant enhancement in pass@k against various baselines. 
Compared with the standalone DeepSeek-Coder, our method Top Pass improves pass@1 from 17.3\% to 20.4\% and pass@5 from 30.7\% to 32.0\%. 
When compared with the previous CodeRanker, our method achieves relative 6.8\% improvement for pass@1 and 2.6\% improvement for pass@5, outperforming baselines by a significant margin.
\Cref{tab:apps} also illustrates the pass@k improvement across various difficulty levels in the APPS benchmark.

\paragraph{MBPP and HumanEval. }
\begin{table}[!t]
\setcellformat[l]{00.0}{}{00.00}
\newcommand{\rc}{\rowcolor{gray!20}}

\centering
\small
\resizebox{\columnwidth}{!}{
\begin{tabular}{lcccc}
\toprule
\multirow{2}{*}[-2bp]{\textbf{Method}} & \multicolumn{2}{c}{\textbf{MBPP}} & \multicolumn{2}{c}{\textbf{HumanEval}}\\
\cmidrule(lr){2-3} \cmidrule(lr){4-5}
                  & \textbf{Pass@1}& \textbf{Pass@5}& \textbf{Pass@1}& \textbf{Pass@5} \\ \midrule
CodeLLaMa         & \msc{47.0}     & \msc{-}        & \msc{36.0}     & \msc{-} \\
WizardCoder       & \msc{51.8}     & \msc{-}        & \msc{57.3}     & \msc{-} \\
code-davinci-002  & \msc{58.1}     & \msc{-}        & \msc{47.0}     & \msc{-}\\
CodeT             & \msc{61.9}     & \msc{-}        & \msc{50.2}     & \msc{-}\\ \midrule
DeepSeek-Coder& \msc{68.7}     & \msb{81.0}     & \msc{60.0}     & \msc{74.5}\\
CodeRanker        & \msc{67.6}     & \msc{77.0}     & \msc{62.8}     & \msc{73.2} \\
Coder-Reviewer     & \msc{66.2}     & \msc{-}     & \msc{62.5}     & \msc{-} \\
\rc Top Pass      & \msb{69.2}     & \msc{79.6}     & \msb{64.6}     & \msb{76.2} \\
\bottomrule
\end{tabular}
}
\caption{Experiment results on MBPP and HumanEval.}
\label{tab:mbpp_humaneval}
\end{table}

\begin{table}[!t]
\setcellformat[l]{00.0}{}{000.0}

\centering
\small
\resizebox{\columnwidth}{!}{
\begin{tabular}{lllll}
\toprule
\textbf{Method} & \textbf{Pass@1} &  \textbf{Pass@2} & \textbf{Pass@3} & \textbf{Pass@5} \\
\midrule
Top Pass & \msc{9.7} & \msc{12.7} & \msc{13.3} & \msc{14.5} \\
Top Pass w/o $L_{pass@k}$ & \msc{7.3} & \msc{9.1} & \msc{10.3} & \msc{13.3} \\
\bottomrule
\end{tabular}
}

\caption{Ablation study on $L_{pass@k}$ on CodeContests.}
\label{table:ablation_passk}
\end{table}
We also conduct experiments on MBPP and HumanEval dataset.
Experiment results are shown in \cref{tab:mbpp_humaneval}. 
On HumanEval dataset Top Pass outperforms both standalone DeepSeek-Coder and CodeRanker, achieving a remarkable 64.6\% pass@1 and 76.2\% pass@5. 
On MBPP dataset Top Pass achieves 69.2\% pass@1, outperforming both standalone code LLMs and CodeRanker. 
It also outperforms other code ranking method such as Normalized Coder-Reviewer with Codex002. 

\subsection{Discussions}
\paragraph{Ablation study.} 
As illustrated in \cref{table:ablation_passk}, the removal of pass@k loss leads to a substantial performance downgrade in pass@k across various values of k.
This observation not only validates the significance of pass@k loss in our method but also highlights its contribution to overall performance.

\begin{figure}[h]
    \centering
    \includegraphics[width=0.96\columnwidth]{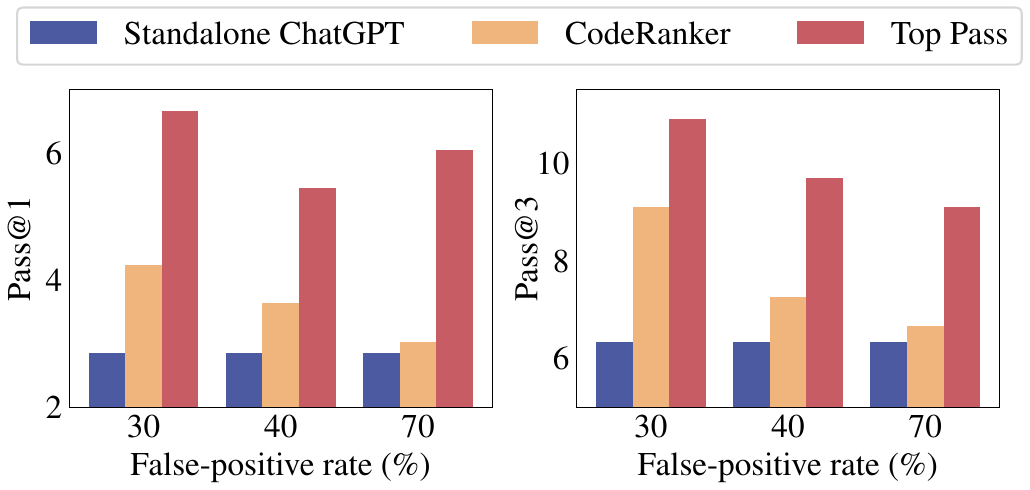}
    \caption{The influence of the false positive rate in the training dataset on various methods, observed through the metric pass@k, where $k=1,3$.}
    \label{fig:fp-cc}
\end{figure}

\paragraph{Robustness to the false positives.} 
Evaluation of the code highly depends on the quality of test cases. However, crafting high-quality test cases demands meticulous consideration.
The insufficient test cases result in labeling buggy code as correct, commonly known as the false positives, in training datasets. Previous study has shown that the existence of such false positives could adversely affect the training of the model~\cite{Li2022}.
To demonstrate the robustness of Top Pass against the false positives, we introduces additional false positive samples by removing different percentage of test cases, yielding three training datasets with distinct false-positive rates: 30\%, 40\%, and 70\%. 
We then evaluate ranker models trained on three generate dataset.
As shown in \cref{fig:fp-cc}, the ranking quality of Top Pass remains minimally effected while pass@k metrics of CodeRanker dramatically drop to slightly better than standalone ChatGPT. This observation aligns with our expectation as Top Pass is robust to low quality code by design.

\begin{figure}[!h]
    \begin{minipage}[b]{0.45\columnwidth}
        \centering
        \includegraphics[width=\columnwidth,height=0.75\columnwidth]{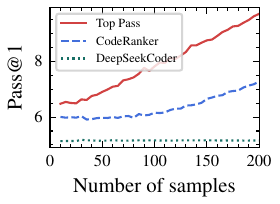}
        \caption{The impact on pass@1 of different sample numbers during test.}
        \label{fig:subfig1}
    \end{minipage}
    \hfill
    \begin{minipage}[b]{0.45\columnwidth}
        \centering
        \includegraphics[width=\columnwidth,height=0.75\columnwidth]{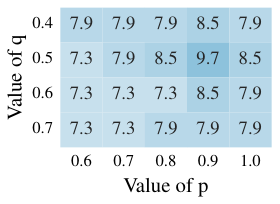} 
        \caption{The impact on pass@1 of different hyper-parameter $p$ and $q$.}
        \label{fig:subfig2}
    \end{minipage}
\end{figure}

\paragraph{Impact of sampling number.} 
We consider the impact of different sample numbers during test and \cref{fig:subfig1} illustrates the change in pass@1 with an increasing number of samples. As depicted in the figure, Top Pass consistently outperforms CodeRanker across all sample numbers, with the lead in pass@1 metric still continuing to increase.
Previous ranking method, CodeRanker, ignores the ranking quality at the top of candidate lists. So when the candidate programs get more and more, the negative programs overwhelm the positive ones at the top, thus leading to the poor pass@1 performance. 
However, our method is dedicated to improve the ranking quality of the top section and achieves outstanding pass@1 performance. 

\paragraph{Impact of hyper-parameters.} 
We also analyze the impact of hyper-parameters $p$ and $q$ by training ranker model on CodeContests dataset with different values of $p$ and $q$. 
From \cref{fig:subfig2} we observe that the variations of two hyper-parameters influence pass@1 smoothly. 
And it is evident that removing bottom negatives brings better performance, which demonstrates that giving more significance to the top positive/negative programs is beneficial. More detailed results are presented in Appendix.

\section{Conclusion}
In this paper, we propose an approach named Top Pass, which elaborates to improve the ranking quality at the top section of the candidate list by designing and directly optimizing a novel pass@k loss. 
Experimental results demonstrate that the Top Pass outperforms previous methods by a substantial margin in term of pass@k metrics across various LLMs and code generation benchmarks with different difficulty levels. 

Our approach improves the usability of the code generation systems and allows users to find correct code within a few tries, thus reducing the burden of handful testing and reviewing. Applying pass@k loss in training reward model in the reinforcement learning process thus improving code LLMs' capacity will be investigated in the future. 

\section*{Acknowledgments}
This research was supported by NSFC (62076121, 61921006) and the Major Program (JD) of Hubei Province (2023BAA024).
The authors would like to thank Hao-Yuan He and Hui Sun for helpful discussion.  


\section*{Appendixes}

\subsection*{Case Study}
Here we analyze an example online judge programming task of CodeContests in detail. 
For CodeForces 1619B task, the problem description is listed as following. 

After sampling 200 candidate programs, the program ranked as the top choice by Top Pass is displayed in \cref{table:topaz}. We can see that the program not only successfully passes all the test cases in CodeForces but also implements the algorithm in a efficient and elegent code style. 
The program ranked as the top choice by CodeRanker is also displayed in \cref{table:coderanker}. The program fails for test cases in CodeForces, thus it is viewed as buggy program. It is not only difficult to read and but also implemented with inefficient algorithm. 

\begin{lstlisting}[style=academic, basicstyle=\ttfamily\tiny, caption={Task description of CodeForces 1619B.}]
Polycarp likes squares and cubes of positive integers. Here is the beginning of the sequence of numbers he likes: 1, 4, 8, 9, .... For a given number n, count the number of integers from 1 to n that Polycarp likes. In other words, find the number of such x that x is a square of a positive integer number or a cube of a positive integer number (or both a square and a cube simultaneously). Input. The first line contains an integer t - the number of test cases. Then t lines contain the test cases, one per line. Each of the lines contains one integer n. Output. For each test case, print the answer you are looking for - the number of integers from 1 to n that Polycarp likes.
\end{lstlisting}

\subsection*{Influence of Sampling Temperature}
\begin{figure}[t]
    \centering
    \includegraphics[width=0.6\columnwidth]{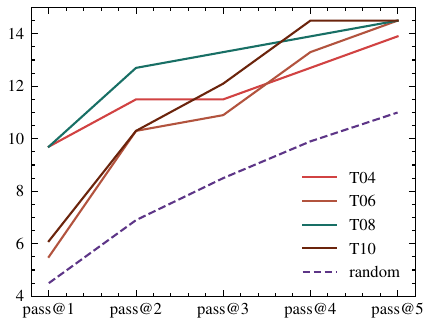}
    \caption{The figure reflects the influence of different sampling temperature on pass@k. In the figure T04, T06, T08, T10 represents the sampling temperature is 0.4, 0.6, 0.8, 1.0. }
    \label{fig:temperature}
\end{figure}

We also investigate the influence of different sampling temperature. The results is displayed in \cref{fig:temperature}. 
From the figure we can see that our method, Top Pass, exhibits stable performance under different sampling temperatures, outperforming random case by a large margin.

\subsection*{Influence of $\lambda$}
We consider the influence of the hyper-parameter  $\lambda$. \Cref{table:lambda} shows the detailed results. From the table we see that different choices of hyper-parameter $\lambda$ has smooth influence on the performance of Top Pass ranker, peaking at 0.3 approximately. 

\begin{table}[h]
\resizebox{1.0\columnwidth}{!}{
\setcellformat[l]{00.0}{}{000.0}

\centering
\small
\resizebox{\columnwidth}{!}{
\begin{tabular}{llllll}
\toprule
\textbf{$\lambda$}  & \textbf{Pass@1} &  \textbf{Pass@2} & \textbf{Pass@3} & \textbf{Pass@5} & \textbf{Pass@10}\\
\midrule
0.0 & \msc{7.9} & \msc{8.5} & \msc{10.9} & \msc{13.9}  & \msc{18.8}\\
0.1 & \msc{7.3} & \msc{9.7} & \msc{11.5} & \msc{13.3}  & \msc{17.0}\\
0.2 & \msc{7.9} & \msc{9.7} & \msc{13.3} & \msc{15.8}  & \msc{18.8}\\
0.3 & \msc{9.7} & \msc{12.7} & \msc{13.3} & \msc{14.5}  & \msc{18.2}\\
0.4 & \msc{8.5} & \msc{9.7} & \msc{13.3} & \msc{15.8}  & \msc{18.2}\\

\bottomrule
\end{tabular}
}

    }
\caption{Influence of different $\lambda$ on pass@k}
\label{table:lambda}
\end{table}

\subsection*{Influence of Different Values of $p$ and $q$}
We train ranker model under different choices of $p$ and $q$. Here we list the performance of pass@k (k=1,2,3,5,10) under different $p$ and $q$ in detail. 
From \cref{table:pq}, we observe that the variations of two hyper-parameters influence performance smoothly as we have declared in the experiments section.

\begin{table}[h]
\resizebox{1.0\columnwidth}{!}{
\setcellformat[l]{00.0}{}{000.0}

\centering
\small
\resizebox{\columnwidth}{!}{
\begin{tabular}{ll|lllll}
\toprule
\textbf{p} & \textbf{q} & \textbf{Pass@1} &  \textbf{Pass@2} & \textbf{Pass@3} & \textbf{Pass@5} & \textbf{Pass@10}\\
\midrule
\multirow{4}{*}{0.6} & 0.4 & \msc{7.9} & \msc{13.9} & \msc{15.2} & \msc{15.8} & \msc{18.2}   \\
 & 0.5 & \msc{7.3} & \msc{12.7} & \msc{15.2} & \msc{15.8} & \msc{17.6}   \\
 & 0.6 & \msc{7.3} & \msc{9.7} & \msc{11.5} & \msc{13.3} & \msc{15.8}   \\
 & 0.7 & \msc{7.3} & \msc{10.9} & \msc{12.1} & \msc{15.2} & \msc{17.0}   \\
\midrule

\multirow{4}{*}{0.7} & 0.4 & \msc{7.9} & \msc{12.1} & \msc{13.3} & \msc{16.4} & \msc{19.4}   \\
 & 0.5 & \msc{7.9} & \msc{10.9} & \msc{12.1} & \msc{13.9} & \msc{17.6}   \\
 & 0.6 & \msc{7.3} & \msc{8.5} & \msc{12.1} & \msc{15.8} & \msc{17.6}   \\
 & 0.7 & \msc{7.3} & \msc{10.9} & \msc{12.1} & \msc{12.7} & \msc{17.0}   \\
\midrule

\multirow{4}{*}{0.8} & 0.4 & \msc{7.9} & \msc{8.5} & \msc{11.5} & \msc{16.4} & \msc{17.0}   \\
 & 0.5 & \msc{8.5} & \msc{10.3} & \msc{12.7} & \msc{16.4} & \msc{19.4}   \\
 & 0.6 & \msc{7.3} & \msc{9.1} & \msc{10.9} & \msc{12.7} & \msc{16.4}   \\
 & 0.7 & \msc{7.9} & \msc{9.1} & \msc{10.9} & \msc{14.5} & \msc{19.4}   \\
\midrule

\multirow{4}{*}{0.9} & 0.4 & \msc{8.5} & \msc{10.3} & \msc{12.7} & \msc{15.8} & \msc{17.6}   \\
 & 0.5 & \msc{9.7} & \msc{12.7} & \msc{13.3} & \msc{14.5} & \msc{18.2}   \\
 & 0.6 & \msc{8.5} & \msc{10.3} & \msc{12.1} & \msc{15.2} & \msc{17.0}   \\
 & 0.7 & \msc{7.9} & \msc{9.1} & \msc{10.9} & \msc{13.3} & \msc{16.4}   \\
\midrule

\multirow{4}{*}{1.0} & 0.4 & \msc{7.9} & \msc{9.7}  & \msc{10.3} & \msc{13.9}  & \msc{17.6}    \\
 & 0.5 & \msc{8.5} & \msc{9.1}  & \msc{10.3} & \msc{12.1}  & \msc{13.9}    \\
 & 0.6 & \msc{7.9} & \msc{9.1}  & \msc{10.3} & \msc{11.5}  & \msc{17.0}    \\
 & 0.7 & \msc{7.9} &\msc{9.1}  & \msc{10.3} & \msc{13.9}  & \msc{17.0}    \\

\bottomrule
\end{tabular}
}

}
\caption{Influence of different values of hyper-parameters $p$ and $q$ on pass@k.}
\label{table:pq}
\end{table}

\newpage
\begin{table*}[!ht]
    \centering
    \footnotesize
    \begin{tabular}{p{\textwidth}}
    \toprule
    Top Pass  \gt{CORRECT} \\
    \midrule 
    \textbf{Program:} 
    \begin{lstlisting}[style=pythonstyle]
import math

def count_special_numbers(n):
    special_numbers = set()

    for i in range(1, math.isqrt(n) + 1):
        sq = i * i
        special_numbers.add(sq)
        cb = sq * i
        if cb <= n:
            special_numbers.add(cb)

    return len(special_numbers)

t = int(input())

for _ in range(t):
    n = int(input())
    print(count_special_numbers(n))
    \end{lstlisting}
    \\
    \bottomrule
    \end{tabular}
    \caption{The program ranked as the top choice by Top Pass. 
    }
    \label{table:topaz}
\end{table*}

\begin{table*}[!ht]
    \centering
    \footnotesize
    \begin{tabular}{p{\textwidth}}
    \toprule
    CodeRanker  \rt{WRONG} \\
    \midrule 
    \textbf{Program:} 
    \begin{lstlisting}[style=pythonstyle]
import math

def count_likes(n):
    cnt = 0
    i = 1
    while i * i <= n:
        cnt += 1
        j = i * i
        while j <= n:
            cnt += 1
            j *= i
            if j <= n:
                cnt -= 1
        i += 1
    return cnt

t = int(input())
for _ in range(t):
    n = int(input())
    print(count_likes(n))
    \end{lstlisting}
    \\
    \bottomrule
    \end{tabular}
    \caption{The program ranked as the top choice by CodeRanker. 
    }
    \label{table:coderanker}
\end{table*}

\clearpage
\bibliographystyle{fcs}
\bibliography{topaz}

\newpage
\begin{wrapfigure}{l}{0.35\columnwidth}
    \centering
    \includegraphics[width=\linewidth]{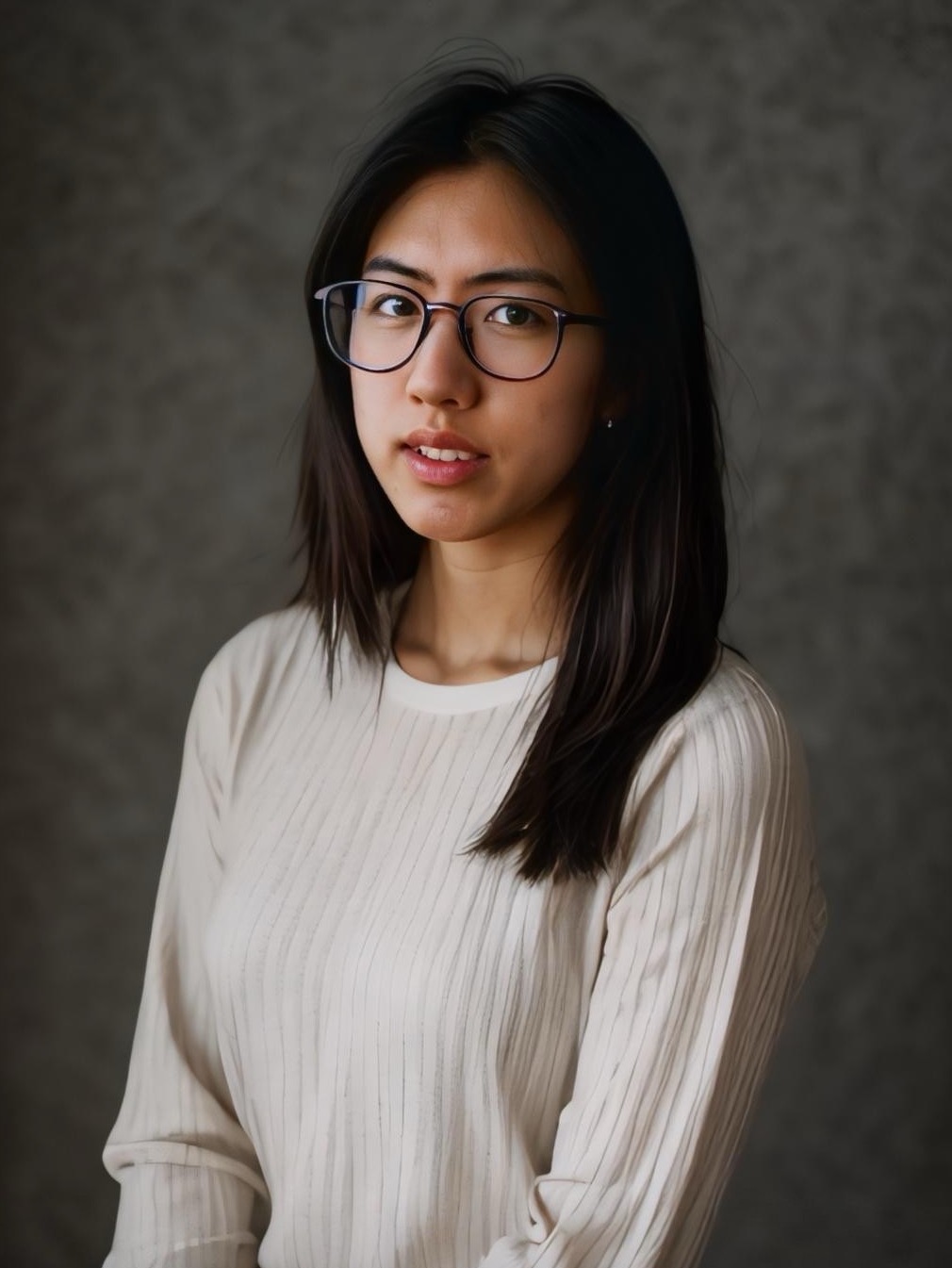}
\end{wrapfigure}

Zhi-Cun Lyu obtained the BSc degree from Nanjing University in 2022. Currently, she is working toward the master degree with the School of Artificial Intelligence, Nanjing University. She is a member of the LAMDA Group. Her research interests mainly include machine learning and data mining, especially in software mining.

\vspace{5pt}

\begin{wrapfigure}{l}{0.35\columnwidth}
    \centering
    \includegraphics[width=\linewidth]{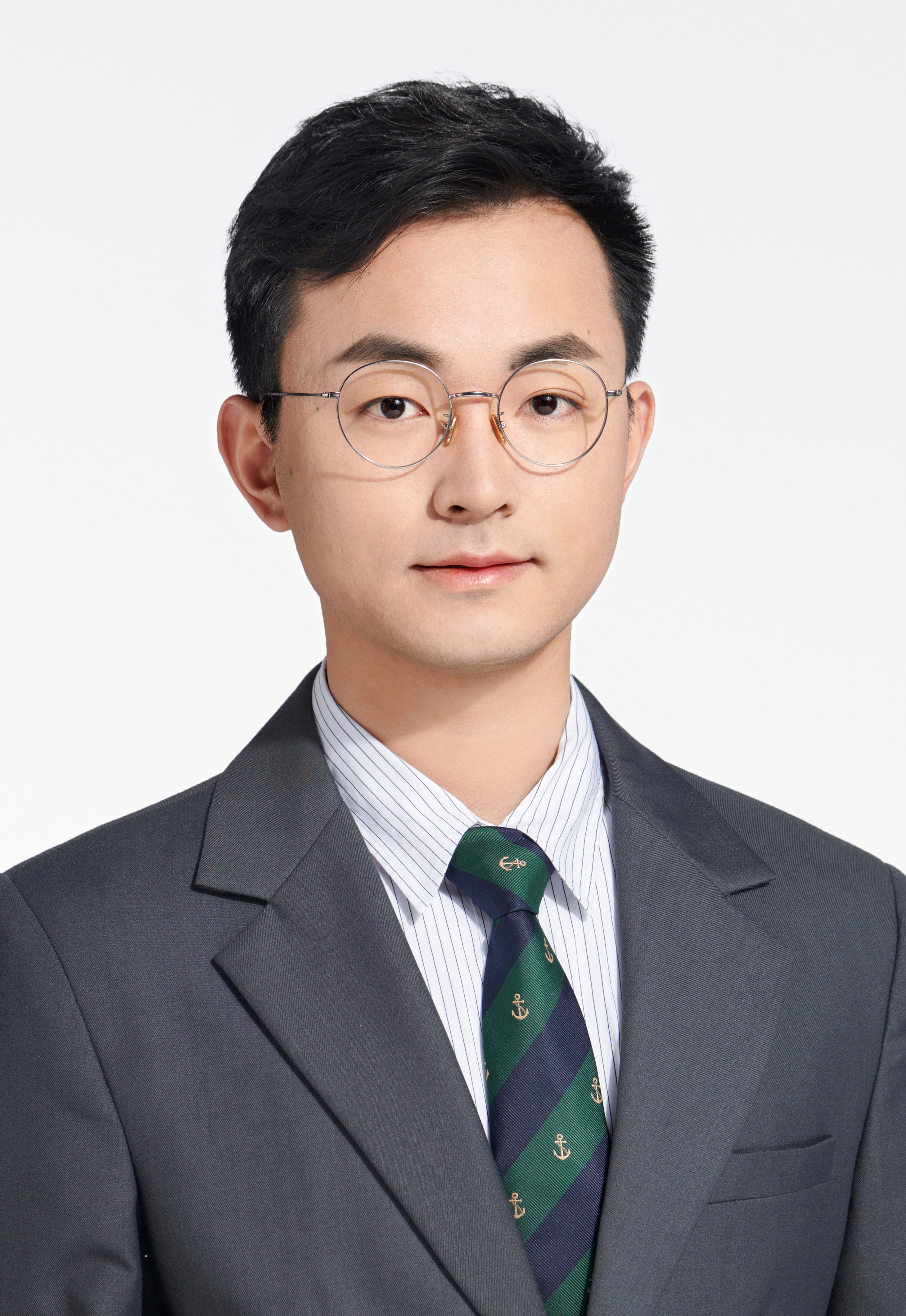}
\end{wrapfigure}

Xin-Ye Li obtained the BSc and MSc degree from Nanjing University in 2020 and 2023 respectively. Currently, he is working toward the PhD degree with the School of Artificial Intelligence, Nanjing University. He is a member of the LAMDA Group. His research interests mainly include machine learning and data mining, especially in software mining. He has received a number of awards including National Scholarship, Outstanding Graduate of Nanjing University and so forth. 

\vspace{5pt}

\begin{wrapfigure}{l}{0.35\columnwidth}
    \centering
    \includegraphics[width=\linewidth]{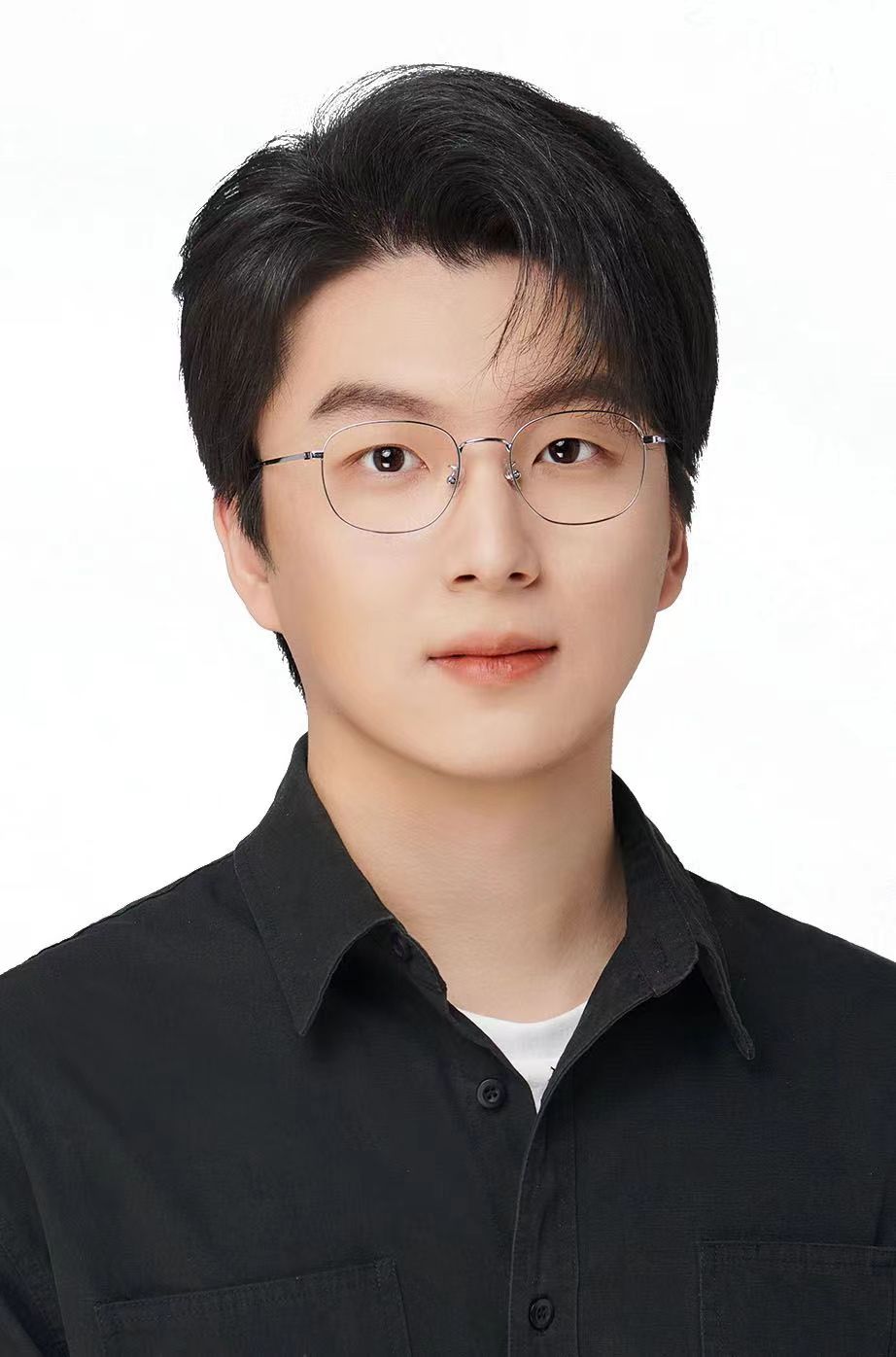}
\end{wrapfigure}

Zheng Xie obtained the PhD degree from Nanjing University in 2023 and the BEng degree from Xi’an Jiaotong University in 2016. Currently, he is working as a researcher at Huawei. His research interests mainly include machine learning and data mining. He has published over 10 papers in top-tier international journals or conference proceedings, including IEEE TPAMI, FCS, AAAI, IJCAI, ICML and so forth.

\vspace{5pt}

\begin{wrapfigure}{l}{0.35\columnwidth}
    \centering
    \includegraphics[width=\linewidth]{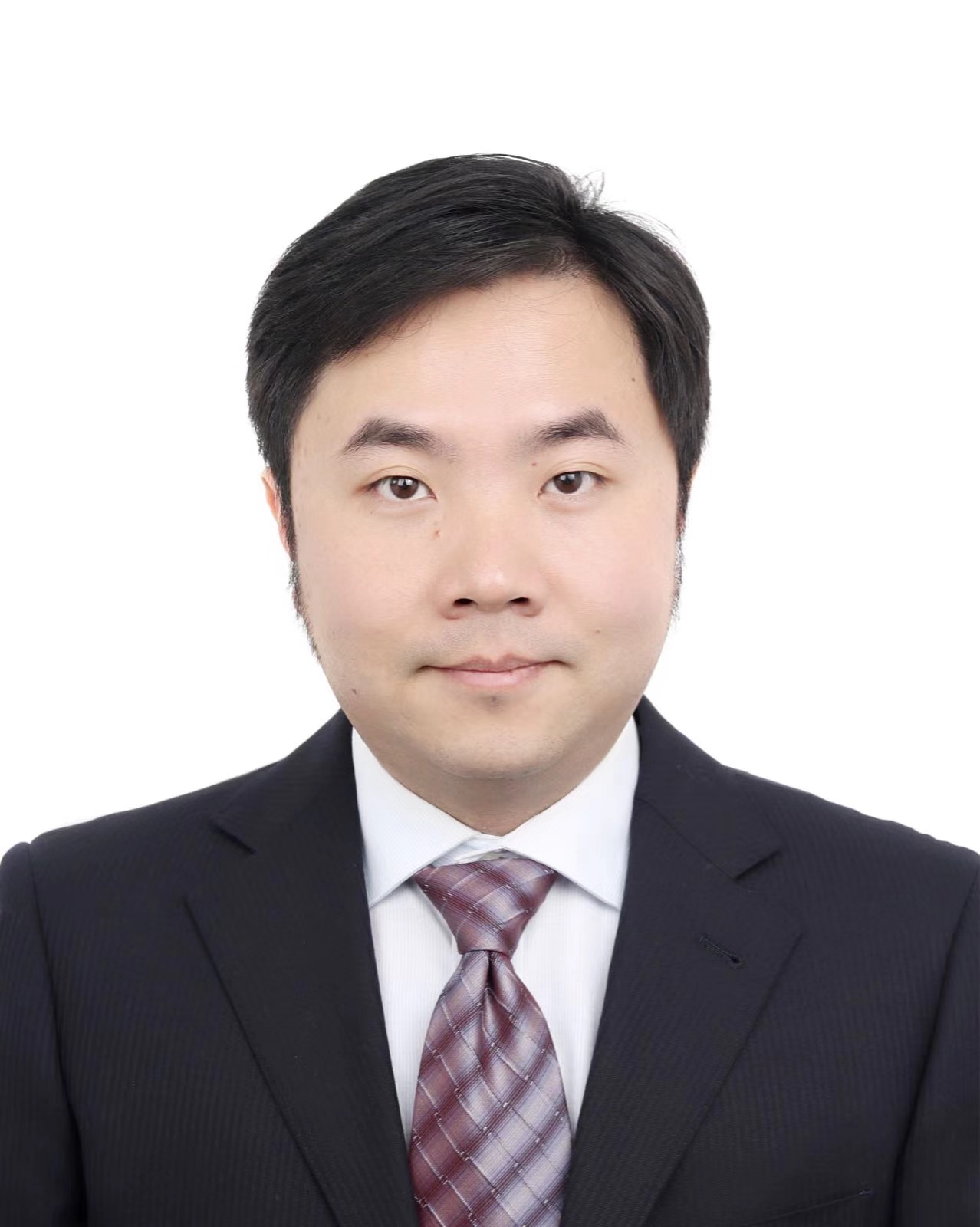}
\end{wrapfigure}
Ming Li is currently a professor at School of Artificial Intelligence, Nanjing University. He is also a member of LAMDA group. His major research interests include machine learning and data mining, especially in software mining. He has served as the area chair of IJCAI, IEEE ICDM, etc, senior PC member of the premium conferences in artificial intelligence such as AAAI, and PC members for other premium conferences, such as KDD, NeurIPS, ICML, etc. He is the founding chair of the International Workshop on Software Mining. He has been granted various awards including the PAKDD Early Career Award, the NSFC Excellent Youth Award, the New Century Excellent Talents program of the Education Ministry of China, etc.
\end{document}